\documentclass{article}
\usepackage{iclr2026_conference,times}
\iclrfinalcopy

\newcommand{\mat}[1]{
    \mathbf{#1}
}

\newcommand{\R}[0]{
    \mathbb{R}
}

\newcommand{\order}[0]{
    \mathcal{O}
}

\usepackage{xspace}

\usepackage{microtype}
\usepackage{amsmath}

\usepackage{amssymb}
\usepackage{gensymb}
\usepackage{graphicx}
\usepackage{pgf}
\usepackage{float}
\usepackage{ifdraft}
\usepackage[hyphens]{url}
\usepackage{hyperref}
\usepackage{enumitem}

\usepackage{eqexpl}
\eqexplSetIntro{where:} 
\eqexplSetDelim{=} 

\usepackage{multicol}

\usepackage{siunitx}
\usepackage{tabularx}
\usepackage{booktabs}
\usepackage{multirow}
\usepackage{tablefootnote}
\usepackage{tabularray}
\UseTblrLibrary{booktabs}
\UseTblrLibrary{counter}

\usepackage{tikz}
\usetikzlibrary{positioning}
\usetikzlibrary{shapes.geometric}
\usetikzlibrary{shapes.arrows}
\usetikzlibrary{decorations.pathmorphing, decorations.pathreplacing, calc}

\usepackage{tikz-cd}
\tikzcdset{
    math mode=false
}

\usepackage{xcolor}

\definecolor{myorange}{RGB}{255, 165, 0}
\definecolor{mydarkseagreen}{RGB}{143, 188, 143}
\definecolor{mydodgerblue}{RGB}{30, 144, 255}

\usepackage{titlesec}

\usepackage{caption}
\usepackage{adjustbox}

\usepackage[titletoc,title]{appendix}

\usepackage{svg}

\usepackage{subcaption}

\usepackage{afterpage}
\usepackage[section]{placeins}

\usepackage{nth}

\usepackage{xstring}

\usepackage{authblk}

\usepackage{todonotes}
\setuptodonotes{inline,backgroundcolor=yellow!20,prepend,caption={TODO}}

\title{NaN-Propagation: A Novel Method for Sparsity Detection in Black-Box Computational Functions}

\author{Peter Sharpe}
\affil{Massachusetts Institute of Technology}
\date{\today}

\begin{document}
\maketitle

\begin{abstract}

    When numerically evaluating a function's gradient, sparsity detection can enable substantial computational speedups through Jacobian coloring and compression. However, sparsity detection techniques for black-box functions are limited, and existing finite-difference-based methods suffer from false negatives due to coincidental zero gradients. These false negatives can silently corrupt gradient calculations, leading to difficult-to-diagnose errors. We introduce NaN-propagation, which exploits the universal contamination property of IEEE 754 Not-a-Number values to trace input-output dependencies through floating-point numerical computations. By systematically contaminating inputs with NaN and observing which outputs become NaN, the method reconstructs conservative sparsity patterns that eliminate a major source of false negatives. We demonstrate this approach on an aerospace wing weight model, achieving a 1.52× speedup while uncovering dozens of dependencies missed by conventional methods -- a significant practical improvement since gradient computation is often the bottleneck in optimization workflows. The technique leverages IEEE 754 compliance to work across programming languages and math libraries without requiring modifications to existing black-box codes. Furthermore, advanced strategies such as NaN payload encoding via direct bit manipulation enable faster-than-linear time complexity, yielding speed improvements over existing black-box sparsity detection methods. Practical algorithms are also proposed to mitigate challenges from branching code execution common in engineering applications.

\end{abstract}

\begin{center}
    \small\textit{Adapted for independent reading from a chapter of the author's Ph.D. thesis.}
\end{center}

\section{Introduction}

Modern engineering design increasingly relies on multidisciplinary design optimization (MDO) frameworks that integrate diverse and complex computational models, as outlined by \cite{martinsEngineeringDesignOptimization2021}. Typically, these frameworks use gradient-based optimization, since gradient-free methods are computationally intractable as problem dimensionality increases due to the curse of dimensionality. In these gradient-based optimization workflows, the computational bottleneck is almost always in computing the gradients of both the objective function and constraint equations.

Many recent frameworks leverage advanced computational techniques such as automatic differentiation (AD) for this gradient computation, as it offers fundamentally better scaling (and accuracy) compared to traditional finite-difference methods as shown by \cite{griewankAutomaticDifferentiation1988,rackauckasEngineeringTradeOffsAutomatic2021}. This enables efficient navigation of high-dimensional design spaces that would otherwise be intractable with traditional design approaches, as demonstrated by \cite{sharpeAcceleratingPracticalEngineering2024,sharpeAeroSandboxDifferentiableFramework2021}.

However, a fundamental challenge arises when these MDO frameworks must incorporate legacy or proprietary analysis codes that cannot be modified or instrumented for automatic differentiation. These \emph{black-box functions} -- computational models where internal implementation details are inaccessible -- are quite common in engineering practice, particularly when progressing beyond the conceptual design stage. Examples include commercial computational fluid dynamics (CFD) solvers, finite element analysis (FEA) packages, vendor-supplied component models, and legacy codes written in older programming languages or compiled libraries.

The integration of black-box functions into gradient-based optimization workflows necessitates using techniques such as finite differencing or the complex-step differentiation method of \cite{martinsComplexstepDerivativeApproximation2003}, which do not rely on code introspection. While these methods can provide the necessary gradient information, they incur significant computational overhead: a function with $n$ inputs requires $n+1$ function evaluations (forward differences) or $2n$ evaluations (central differences) to compute its complete Jacobian. For high-fidelity engineering analyses where individual function evaluations may require minutes to hours of computation, and the number of inputs may correspond to thousands of possible design variables, this overhead becomes prohibitive.

\subsection{The Role of Sparsity in Gradient Computation}

The computational burden of black-box gradient evaluation can be substantially reduced through sparsity exploitation. In engineering analysis, most input-output relationships exhibit structural sparsity -- many inputs do not influence certain outputs due to the underlying physics and mathematical formulation. For instance, consider a classical static structural analysis using an Euler-Bernoulli cantilevered beam model as described by \cite{drelaIntegratedSimulationModel1999}. Here, any local forces applied near the beam's root have no influence on the internal stress near the beam's free tip -- the influence dependency is purely one-directional: tip to root.

Knowledge of this sparsity pattern enables \emph{Jacobian compression} through the graph coloring algorithms described by \cite{gebremedhinEfficientComputationSparse2009, gebremedhinWhatColorYour2005}. When we know that certain input variables do not interact in their influence on any output, their gradients can be computed simultaneously using appropriately constructed perturbation vectors. This can reduce the number of required function evaluations from $\mathcal{O}(n)$ to as few as $\mathcal{O}(\chi)$, where $\chi$ is the chromatic number of the sparsity pattern's column intersection graph -- often much smaller than $n$ for engineering problems.

The potential impact of sparsity exploitation is substantial, though problem-dependent. In particularly well-suited cases, like trajectory optimization via direct collocation transcription as described by \cite{kellyIntroductionTrajectoryOptimization2017,kellyTranscriptionMethodsTrajectory2017}, evaluation of the gradient of a function with $n$ inputs can be performed in $\mathcal{O}(1)$ time. While other cases may not see such a dramatic reduction, any improvement in accelerating the bottleneck step of gradient computation is quite helpful.

\subsection{Existing Approaches}
\label{sec:nan-existing-methods}

Currently, the most common approach to obtain the sparsity pattern of a black-box function is a \emph{gradient-based} heuristic method. First, the (dense) Jacobian is constructed via finite differencing, evaluated at some particular point in the input space that is assumed to be representative. (In the context of an optimization framework, this input point usually corresponds to the initial guess.) Then, a critical assumption is made: it is assumed that any zero entries in this evaluated Jacobian are indicative of a zero entry in the true Jacobian -- in other words, sparsity seen at one point indicates sparsity at all points. Due to this assumption, the sparsity pattern constructed with this method is only an estimate, not a guarantee.

The fact that this reconstructed sparsity pattern is only an estimate is inevitable for black-box functions. To illustrate why, consider a piecewise function, where input $x_j$ influences output $f_i$ when $x_j$ is within a certain interval but not outside of it. It is possible to construct a pathological black-box function where this region of influence is arbitrarily small, such that it will nearly never be detected by direct sampling of the Jacobian. Such cases will result in a \emph{false negative} in the estimated sparsity pattern: cases where $x_j$ and $f_i$ are believed to be independent, when in reality they are not.

This kind of branching code execution is perhaps the most common cause of such false negatives (often via conditional expressions) in the context of engineering analysis. However, there are other relevant failure modes as well. One possible cause of false negatives is if the representative input point yields a partial derivative of zero only due to coincidence, rather than due to mathematical structure. One simple illustrative example is the function $f(x) = x^2$, if the evaluation point of $x = 0$ is chosen. Here, a central finite difference will see zero gradient and incorrectly infer no dependency. On most functions of practical interest, such points are usually rare within the input space\footnote{In most practical functions $\vec{f}(\vec{x}): \R^m \rightarrow \R^n$, the dimensionality of all manifolds where $\partial f_i/\partial x_j=0$ is less than the dimensionality of the input $\vec{x}$. In such cases, \emph{almost all} points will be free of coincidental zeros.}. However, in an optimization context, the distribution of user-specified initial guesses is not uniform, and in fact adversarial: users will preferentially supply near-optimal initial guesses, which have near-zero gradients (if the problem is unconstrained). Because of this, and as demonstrated later in Section \ref{sec:nan-comparison}, this effect can contribute to a non-negligible rate of false negatives in practice.

\emph{False positives}, where $x_j$ and $f_i$ are believed to be related when they are not, are also possible, though rarer. A practical example where this could occur is with a black-box function that has truncation error, perhaps due to a nonlinear solver or numerical integrator that uses a value-dependent convergence criterion. In such cases, changing one input may alter the number of iterations required for convergence. This can cause changes in the truncation error of all outputs -- even those that are mathematically-unrelated to that input. Thus, changes in this truncation error (i.e., a differing number of iterations) may yield a nonzero Jacobian entry, even if the math that the code aims to represent has no such dependency. Other false positive scenarios, such as stochastic functions, are possible but relatively uncommon in engineering analysis. In either the false-negative or false-positive case, an interesting observation is that the sparsity of the function \emph{in code} may not match the sparsity of the mathematical model the code aims to represent.

For reasons related to Jacobian compression (described later via demonstration in Section \ref{sec:nan-jacobian-compression}), the harm caused by a false-negative and a false-positive are very unequal -- false negatives are far worse. The ultimate effect of a false-positive is that subsequent gradient calculations will be slightly slower, though still numerically correct. On the other hand, the ultimate effect of a false-negative is that gradient calculations can be incorrect, due to erroneous Jacobian compression. Worse yet, these incorrect results can occur silently. In the context of design optimization, an incorrect gradient might only be detected much later in the development effort, when an optimizer consistently cannot converge (as the KKT conditions cannot be satisfied). By this point in the code development process, the combined optimization model is usually much more complex, and finding and eliminating an incorrect-gradient error becomes enormously tedious. Because of this, a new method for black-box sparsity estimation that eliminates false-negative failure modes offers significant value to the end-user, even if it comes at the cost of false-positive failure modes. In other words, the focus of black-box sparsity estimation within this context should be to build up a \emph{conservative} estimate of the sparsity pattern that trades away specificity in favor of sensitivity.

Recently, \cite{hillSparserBetterFaster2025,rackauckasGeneralizingAutomaticDifferentiation2021} have developed operator-overloading-based approaches to sparsity detection that do not suffer from this local-to-global problem. However, the increased robustness of such methods comes at the cost of generality: an overloading-based sparsity detection framework can only be used on code that uses a supported math library and programming language (in both of the cited examples, Julia). Therefore, these methods cannot currently be applied to general black-box functions, which may be in any language.

\subsection{Contributions and Paper Organization}

This paper's primary contributions are:

\begin{enumerate}[noitemsep]
    \item A novel sparsity detection technique that eliminates the coincidental zero gradient failure mode of existing methods
    \item Theoretical analysis and practical demonstration of the method's advantages and limitations
    \item Advanced algorithms including NaN payload encoding for improved computational complexity
    \item Specialized strategies for engineering analysis functions with branching execution paths
    \item Comprehensive evaluation on a representative aerospace engineering application
\end{enumerate}

The remainder of this paper is organized as follows: Section \ref{sec:nan-demo} introduces the core NaN-propagation technique with a detailed demonstration. Section \ref{sec:speed} analyzes computational performance considerations. Sections \ref{sec:limitations} and \ref{sec:mitigation} examine potential limitations and mitigation strategies. Section \ref{sec:nan-comparison} provides quantitative comparison with existing methods. Section \ref{sec:nan-advanced} presents advanced algorithmic improvements. Section \ref{sec:conclusions} summarizes key findings and discusses future directions.

\section{NaN-Propagation: Method and Demonstration}
\label{sec:nan-demo}

This manuscript introduces a novel technique that we call \emph{NaN-propagation} to trace these input-output dependencies through black-box functions. The name refers to ``not-a-number'' (NaN), which is a special value in floating-point arithmetic that is often used to represent undefined, missing, or otherwise unrepresentable values. The proposed technique exploits the fact that Not-a-Number (NaN) values are universally propagated through floating-point numerical computations; colloquially, they tend to ``contaminate'' any calculation that is given a NaN input. For example, a dyadic\footnote{referring to a function that accepts two input parameters} operation with just one NaN operand will return NaN. This behavior is explicitly part of math library APIs that follow the IEEE 754 standard established in 1985. Because nearly all floating-point math libraries created since the advent of IEEE 754 follow it, this presents a fascinating way to potentially trace sparsity in a way that is essentially independent of math library or programming language. By systematically and deliberately contaminating inputs to a black-box function with NaN values and observing which outputs become NaN, a provably-conservative estimate of the sparsity pattern can be reconstructed.

As a high-level comparison against the existing technique described in this section, this proposed new technique:

\begin{itemize}[noitemsep]
    \item Eliminates a false-negative failure mode, namely coincidental zero gradients.
    \item Adds a new false-positive failure mode, for functions with internal mathematical cancellation.
    \item Is either equivalent in runtime speed, or potentially much faster due to \emph{short-circuiting operators}. This consideration is primarily determined by behavior of the math library used by the black-box function.
    \item Is compatible with most black-box functions, but not all, due to possible internal handling of NaN values. However, this incompatibility is usually easily and immediately detected, allowing the user to fall back to the existing technique.
\end{itemize}

This new technique, and strategies and mitigations around these four high-level tradeoffs, are the focus of the remainder of this work.

\subsection{Example Black-Box Function: Wing Weight Model in TASOPT}

\begin{figure}[H]
    \centering
    \includegraphics[width=4in]{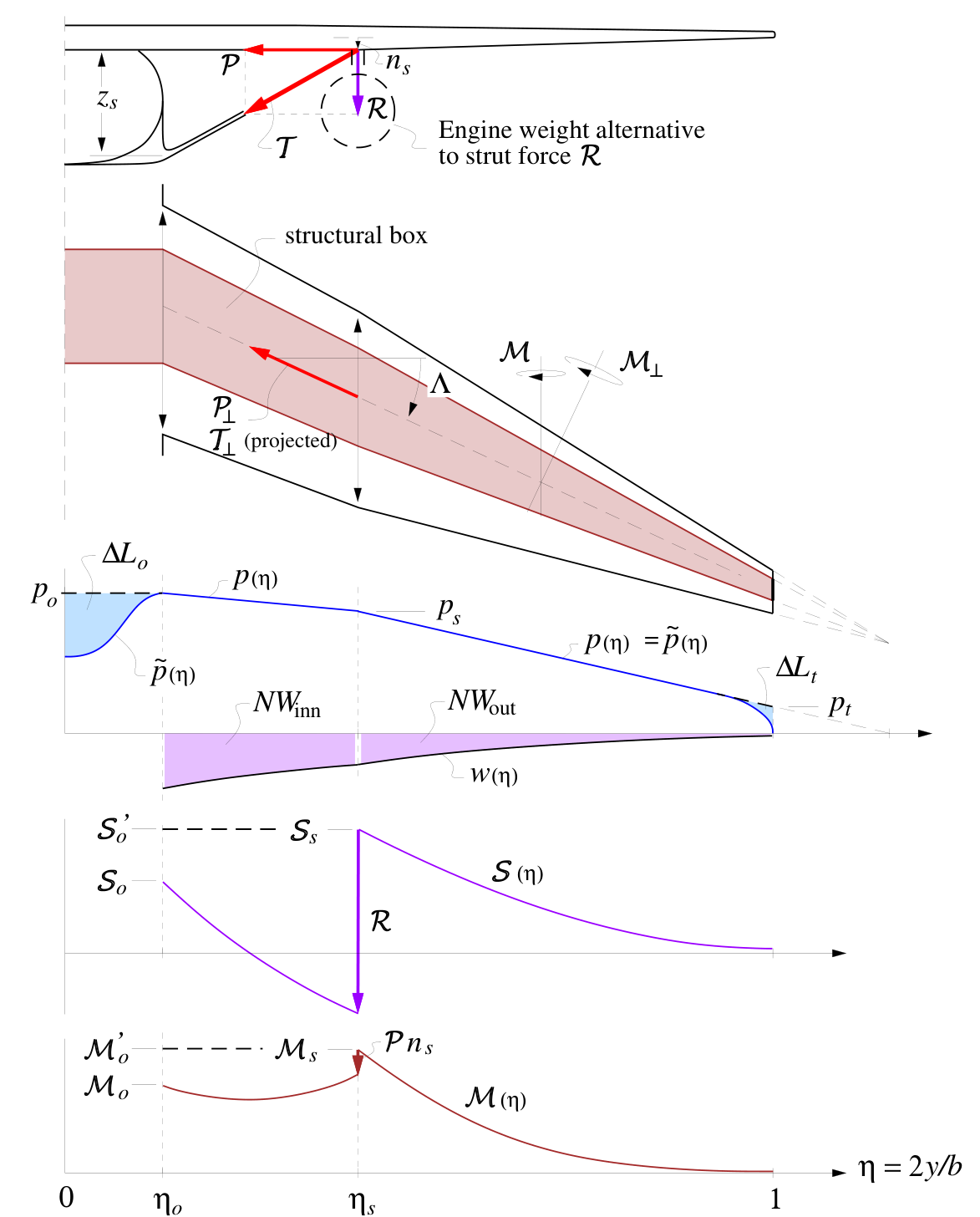}
    \caption{Illustration of TASOPT's wing weight model, reproduced from \cite{drelaTASOPTTransportAircraft2010}. The model uses an Euler-Bernoulli beam model to compute shear and moment distribution, and accepts a relatively large set of geometric parameters as inputs.}
    \label{fig:nan-tasopt-wing-weight}
\end{figure}

Some of the more nuanced details of the proposed NaN-propagation technique are most clearly explained through demonstration. To do this, we leverage one of the constituent submodels within \emph{TASOPT}, an aircraft design optimization suite for tube-and-wing transport aircraft developed by \cite{drelaTASOPTTransportAircraft2010}. The submodel we use is a wing weight model\footnote{In TASOPT, this model is named \texttt{surfw}.}, with the major modeling considerations shown in Figure \ref{fig:nan-tasopt-wing-weight}. The model is relatively detailed, with 38 inputs and 37 outputs. Inputs include geometric parameters, material properties, and configuration-level decisions (e.g., the number and location of engines, or the presence and dimensions of a strut). Outputs are the weights of various wing components, metrics describing the weight distribution, stresses at key points, and other quantities of interest.

For the purposes of this demonstration, the code implementation of this model is accessed through \emph{TASOPT.jl}, which is a transpilation of the original Fortran code into Julia performed by \cite{tasopt_jl}. We aim to estimate the sparsity pattern of this function from within Python while accessing this model only through a Python-Julia interface, making this a true black-box function written in an entirely separate programming language.

\subsection{Sparsity Evaluation}

The first step in the NaN-propagation technique is to obtain a representative input. In the context of an optimization problem, the initial guess provides a natural choice:

\begin{figure}[H]
    \centering
    \includegraphics[page=2, width=\textwidth]{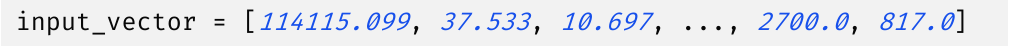}
\end{figure}

Next, we create a set of ``contaminated'' inputs, essentially by merging this input with one-hot encoded NaN values:

\begin{figure}[H]
    \centering
    \includegraphics[page=3, width=\textwidth]{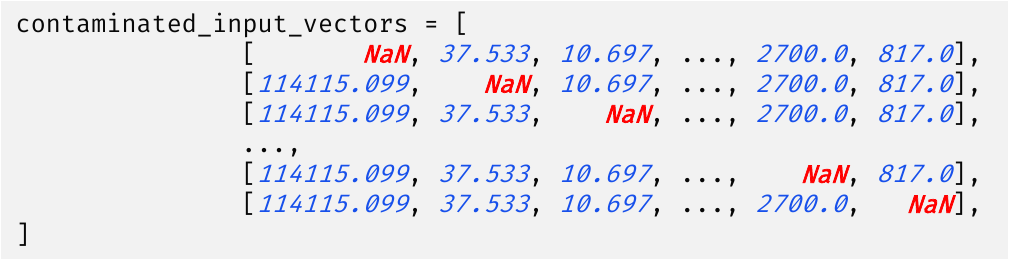}
\end{figure}

Finally, we evaluate the black-box function at each of these contaminated inputs, and observe which outputs become NaN. For example, the output for one such contaminated input is shown below:

\begin{figure}[H]
    \centering
    \includegraphics[page=4, width=\textwidth]{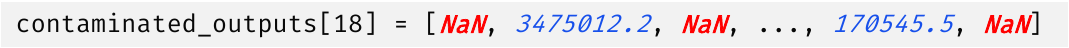}
\end{figure}

\begin{figure}[H]
    \centering
    \includegraphics[width=\textwidth]{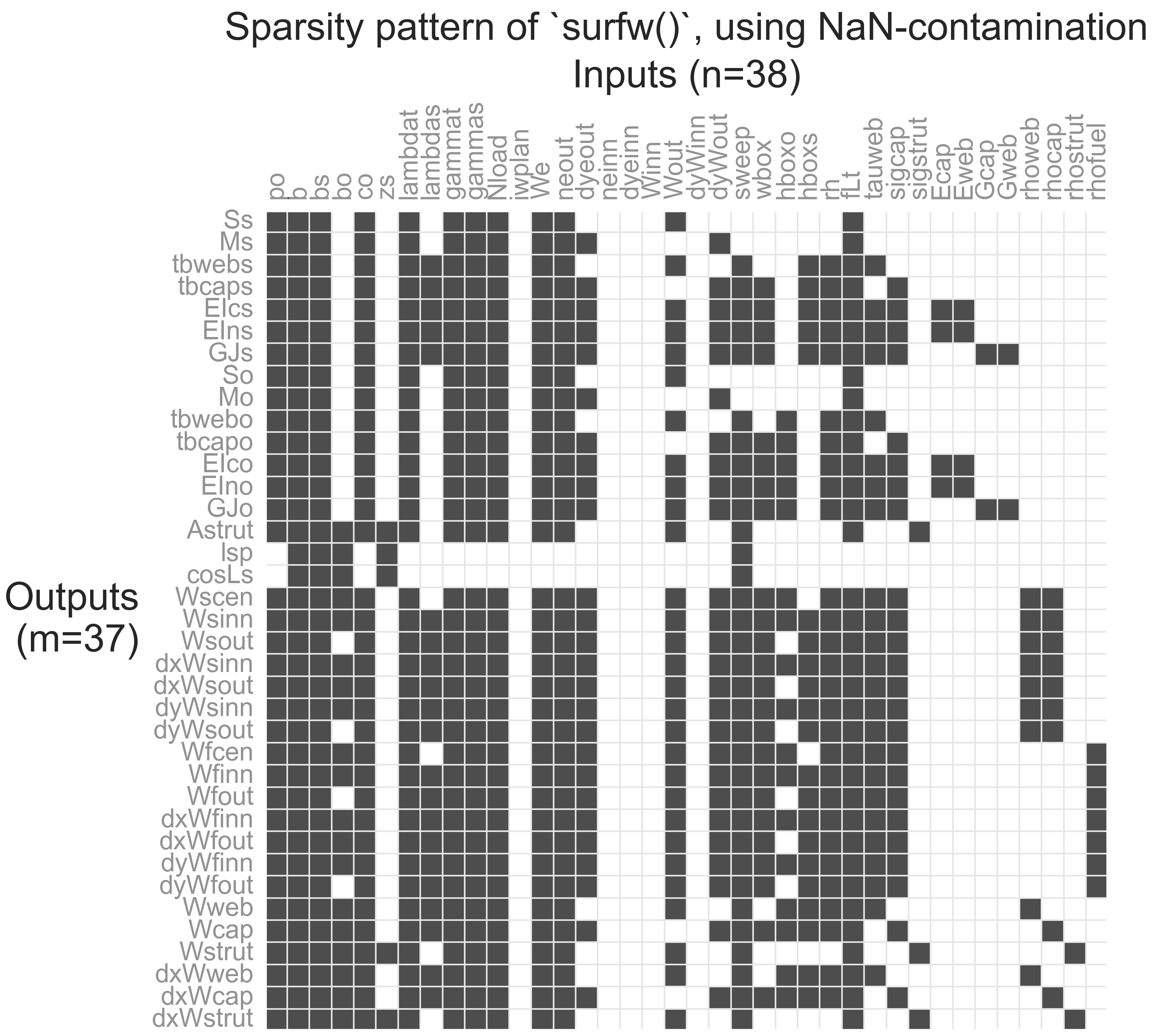}
    \caption{Sparsity pattern of the TASOPT wing weight model, estimated using NaN-propagation. In this visualization, gray squares indicate nonzero entries in the Jacobian, while white squares indicate zero entries.}
    \label{fig:nan-jacobian}
\end{figure}

This simple procedure yields the complete bipartite graph of which inputs and outputs are linked, and this can be interpreted as a sparsity pattern. For the demonstration case study, the sparsity pattern is shown in Figure \ref{fig:nan-jacobian}.

A key result is that this NaN-contamination technique eliminates the possible false-negative failure mode of coincidental zero gradients that can occur with existing gradient-based methods. Consider the example case discussed in Section \ref{sec:nan-existing-methods}, where the sparsity of the function $f(x) = x^2$ is traced at $x=0$. Here, a NaN-contamination technique will still detect the dependency between $x$ and $f(x)$, while existing methods will not. Because the potential consequence of a false-negative is so large, this represents a significant advantage over existing methods.

\subsection{Jacobian Compression}
\label{sec:nan-jacobian-compression}

This sparsity pattern can be used to enable gradient computation accelerations via simultaneous evaluation. More precisely, this involves Jacobian compression across columns, since the gradients of this black-box will later be obtained with finite-differencing (``forward-mode''-like behavior). Briefly, the steps to achieve this are:

\begin{enumerate}
    \item Convert the sparsity pattern into an undirected graph representation where each node corresponds to an input (i.e., a column in Figure \ref{fig:nan-jacobian}), and the presence of each edge indicates that the corresponding pair of inputs has overlapping sparsity. Conveniently, the adjacency matrix of this graph can be obtained by computing the Gramian of the sparsity pattern. If the sparsity pattern is represented as the binary matrix $S$, then the adjacency matrix of the graph is the binary matrix $S^T S \neq 0$.
    \item Apply a vertex coloring algorithm to this graph. This is a well-studied problem with many efficient approximate algorithms available, as discussed by \cite{kubaleGraphColorings2004}.
\end{enumerate}

\begin{figure}[H]
    \centering
    \includegraphics[width=\textwidth]{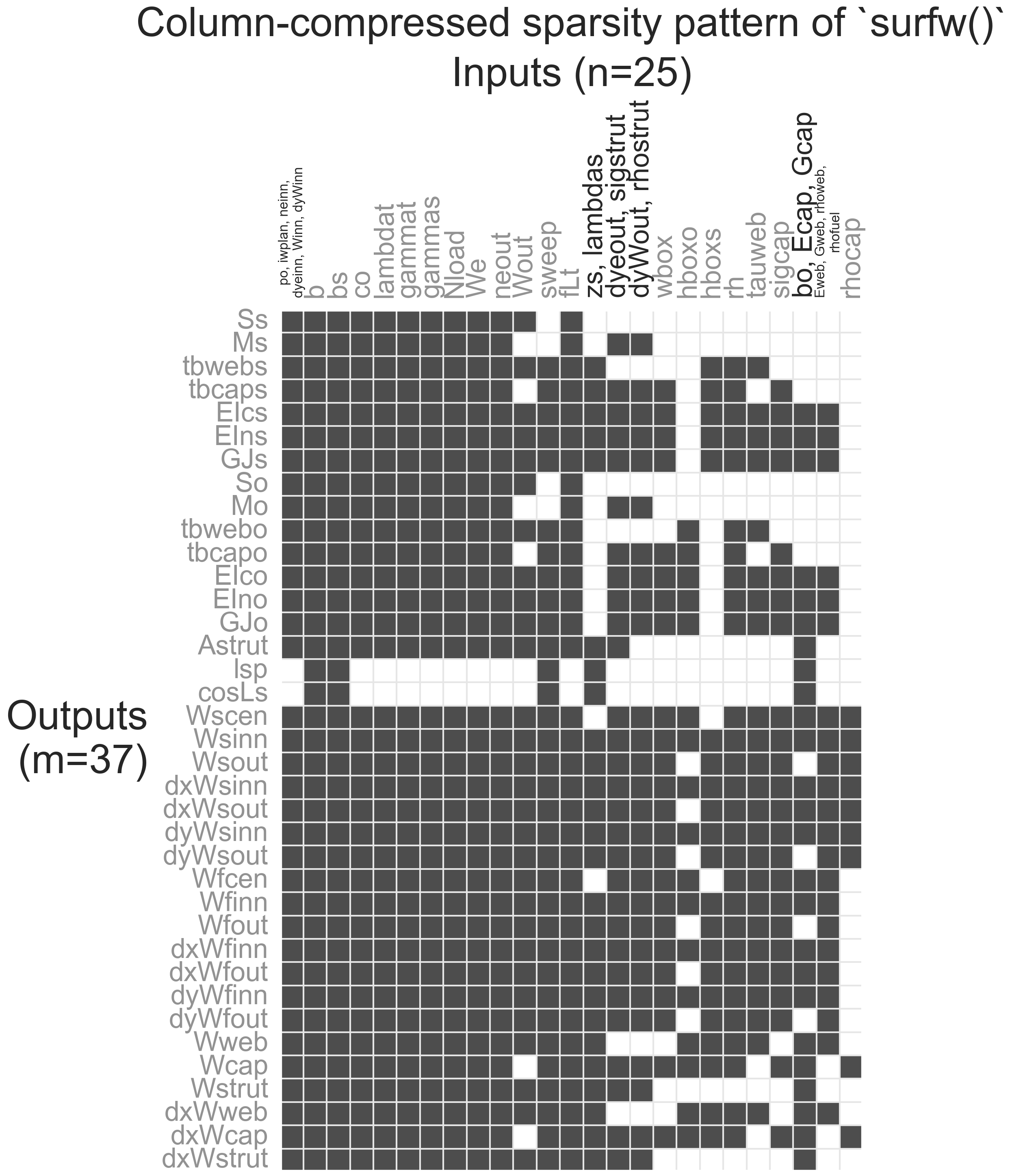}
    \caption{Compressed sparsity pattern of the TASOPT wing weight model, obtained by applying vertex coloring to the sparsity pattern of Figure \ref{fig:nan-jacobian}.}
    \label{fig:nan-jacobian-compressed}
\end{figure}

Visually, the result of this process can be shown with the compressed Jacobian of Figure \ref{fig:nan-jacobian-compressed}. Here, several columns show a combination of multiple inputs, and gradients with respect to these inputs can be safely computed simultaneously due to structural independence. The improvement in gradient computation speed is case-dependent, but in this demonstration, a reduction from 38 inputs to 25 effective inputs is achieved. Because gradient computation runtime scales linearly with the number of inputs in a finite-difference (or complex-step) context, this means that all future gradient calculations are roughly $38/25 \approx 1.52$x faster than a dense Jacobian construction.

\section{Computational Performance Analysis}
\label{sec:speed}

This gradient acceleration comes at a small upfront computational cost, since performing the sparsity trace requires a number of function evaluations equal to the number of inputs. This is true for both existing methods (described in Section \ref{sec:nan-existing-methods}) and for NaN-contamination. This cost is nearly always worthwhile in the context of design optimization: for the TASOPT wing model, a sparsity trace ``pays for itself'' in runtime if the subsequent optimization process requires more than three gradient evaluations.

In some cases, the runtime of a NaN-contamination-based sparsity trace will be essentially equal to that of standard finite-difference-based sparsity estimate. Intuitively, this makes sense, because both approaches require the same number of black-box function evaluations. However, in other cases, NaN-contamination can actually be significantly faster due to operator \emph{short-circuiting}. In this context, short-circuiting refers to operators that will check for NaN input values before performing computation; if any are found, the computation is skipped and a NaN is immediately returned instead. Programming languages and math libraries vary widely in whether and how they implement NaN-short-circuiting, so the magnitude of this potential speed-up is highly case-dependent.

\section{Potential Limitations and Failure Modes}
\label{sec:limitations}

While NaN-contamination eliminates a notable false-negative failure mode compared to existing methods, it also has several possible pitfalls that merit discussion. Some of these pitfalls are shared with existing methods, while others are unique to this technique.

\subsection{Mathematical False-Positives}

One limitation of NaN-contamination is that it can introduce new false-positive failure modes. To illustrate one possible mechanism, consider a scenario where we attempt to NaN-propagate through a black-box function written in code as $f(x)= x - x$. If $x$ is real-valued, then the correct result of a sparsity trace is that the input $x$ and the output $f(x)$ are structurally unrelated. However, a NaN-contamination technique will indicate a possible dependency here, as NaN values do not subtractively cancel. In contrast, existing gradient-based methods will correctly identify independence. More generally, this kind of false-positive can occur in any function where self-cancellation leads to structural independence, such as $f(x) = \sin(x)^2 + \cos(x)^2$ or $f(x) = x^0$.

In some ways, this false-positive is an unavoidable result of the same properties that allow NaN-contamination to eliminate false negatives due to coincidental zero gradients. In theory, global knowledge of this function as a computational graph could allow this self-cancellation to be detected and avoided. However, this would require a level of introspection that is fundamentally incompatible with the black-box nature of the function.

\subsection{Internal NaN Handling}

While NaN-propagation does allow sparsity detection of many black-box functions, including those in other programming languages, there are some cases where it cannot be employed. In particular, black-box functions that internally handle NaN values may not allow propagation. Most commonly in this scenario, the function will instead raise an error and halt the computation. This behavior is easily and immediately detected, and the sparsity detection routine can fall back to existing gradient-based methods in these cases.

A more serious false-negative failure mode can occur if the program actively overwrites NaN outputs without raising an error, as this can result in false negatives in the sparsity trace. Fortunately, this overwriting behavior is quite rare in engineering analysis code today. The main way this can occur is if the black-box code returns special flag values (sometimes called \emph{magic numbers}) to signal invalid computation, though this is generally considered an anti-pattern and avoided in numeric code. Even in code that has this overwriting behavior upon seeing a NaN result, most codes will at least raise a warning to the user.

Another possible false-positive failure mode can occur if operators within the black-box are overly aggressive about propagating NaN values. For example, consider a simple matrix-vector multiplication of the form $\vec{f}(\vec{x}) = \mat{A} \vec{x}$. If a single element of the matrix $A_{ij}$ is NaN, this should result in a single NaN output $f_i$, while the remaining elements are unaffected. Most programming languages and math libraries (e.g., NumPy, Julia) implement this behavior, allowing an accurate sparsity trace. However, some older libraries may aggressively contaminate and yield a NaN output for the entire output vector, which effectively results in false-positives.

\subsection{Branching Code Execution}
\label{sec:nan-bce}

Finally, branching code execution remains a challenge for NaN-contamination-based sparsity tracing, just as it is with existing methods. An example of this can be shown with the wing weight demonstration problem described in Section \ref{sec:nan-demo}. Here, the input variable \texttt{iwplan} is essentially an enumerated type that controls the wing configuration. Specifically:

\begin{itemize}[noitemsep]
    \item \texttt{iwplan = 0} or \texttt{iwplan = 1} corresponds to a cantilever wing without a strut.
    \item All other values of \texttt{iwplan} correspond to a wing with a strut.
\end{itemize}

\begin{figure}[H]
    \centering
    \begin{subfigure}{0.49\textwidth}
        \centering
        \includegraphics[width=\textwidth]{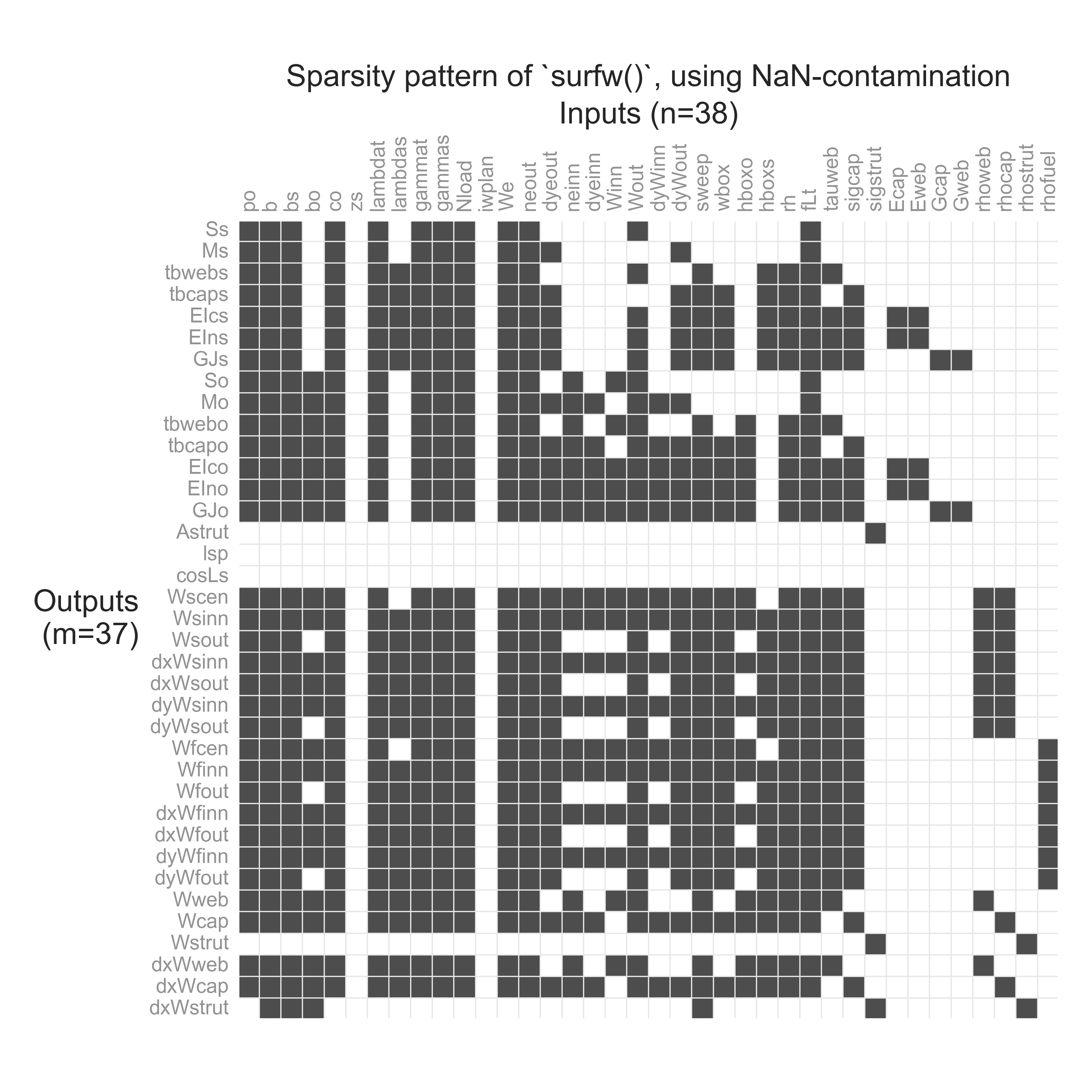}
        \caption{Without strut (\texttt{iwplan = 1})}
    \end{subfigure}
    \begin{subfigure}{0.49\textwidth}
        \centering
        \includegraphics[width=\textwidth]{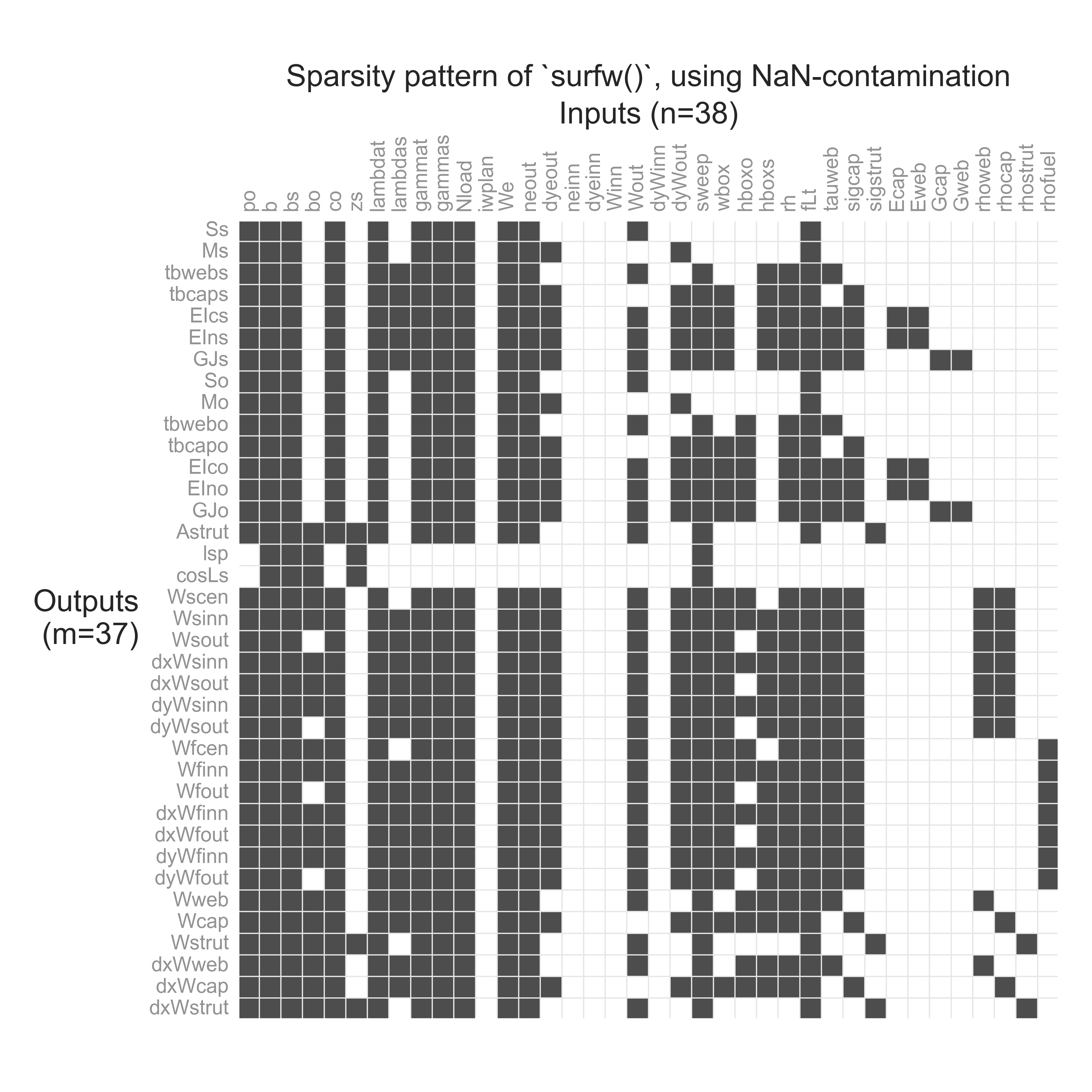}
        \caption{With strut (\texttt{iwplan = 2})}
    \end{subfigure}
    \caption{Sparsity pattern of the TASOPT wing weight model, estimated using NaN-propagation, as the presence of a strut is varied by changing an input variable. Sparsity patterns differ by the presence or absence of interactions across three rows and columns of the matrix.}
    \label{fig:nan-jacobian-branching}
\end{figure}

Perhaps unsurprisingly, the sparsity pattern of the function becomes materially different if the wing configuration is changed by adding a strut. This is shown in Figure \ref{fig:nan-jacobian-branching}.

With respect to branching code execution, NaN-contamination based strategies may have an advantage over existing methods due to the presence of \emph{static conditional} operators. These operators, which are offered in several math libraries, allow for conditional statements to be represented at an \emph{operator} level rather than a \emph{language} level. (In many languages, this is a ternary operator with a name similar to \texttt{ifelse}, \texttt{cond}, or \texttt{where}.) In such operators, both branches of the conditional are evaluated, yet only one is returned. This contrasts with language-level conditionals (e.g., traditional \texttt{if} statements), where the actual path of code execution changes. In some math libraries (e.g., CasADi as described by \cite{anderssonCasADiSoftwareFramework2019}) static conditionals allow NaNs to propagate through regardless of the condition -- if either branch evaluates to NaN, the result is a NaN even if the NaN branch would not ordinarily be taken. Because of this, NaN-contamination-based sparsity tracing can potentially handle branching code execution better than existing options, depending on the math library used by the black box.

\section{Mitigating Branching Code Execution for Black-Box Engineering Analyses}
\label{sec:mitigation}

While branching code execution inherently poses problems for black-box sparsity tracing in the general sense, there are heuristic strategies that can be used to improve performance in practical cases. In particular, we look at strategies that tend to work well specifically in the context of engineering analysis code. In these cases, code branching often corresponds to variables that either change the design configuration or the mode of an analysis. The \texttt{iwplan} variable discussed in Section \ref{sec:nan-bce} is an example of this. With this in mind, we can make two observations specific to engineering analysis code that can be exploited.

First, inputs that trigger branching code execution -- collectively, ``flag inputs'' -- are often represented as discrete data types (e.g., integers, booleans, or strings) rather than continuous types (floating-point values). Because of this, knowledge of the input data types would allow some inferences to be drawn about which inputs are likely to trigger branching execution. In some cases, users may be able to manually mark flag inputs, either by leveraging domain knowledge or by inspecting known type-stable code\footnote{e.g., if the black-box function is written in a statically-typed language, and the user can view at least the function signature}. Of course, in the general case of a black-box function, this ability to inspect a function's type signature (either manually or dynamically) is not guaranteed. However, one possible option is to perform automatic type inference at runtime based on the types that constitute the initial guess.

Secondly, we can also take advantage of the fact that, in many cases, the value of flag inputs remains fixed during a single optimization study. For example, the design configuration may be kept fixed in a sizing study, or an analysis may only be run under certain assumptions. In such cases, the potential damage of a false negative is nullified, because the model is never evaluated on the other code branch.

One possible solution that takes advantage of both of these observations is to use a greedy approach. Concretely:

\begin{enumerate}
    \item Run an initial sparsity trace on the black-box model using the initial guess as the input.
    \item Every time a new value is observed on a discrete-typed input, re-run the sparsity trace.
    \item Take the result of this sparsity trace, and compute the union with all previously-observed sparsity patterns for this black-box model. Use this new sparsity pattern for Jacobian compression in subsequent evaluations.
\end{enumerate}

\begin{figure}[H]
    \centering
    \includegraphics[page=9, width=\textwidth]{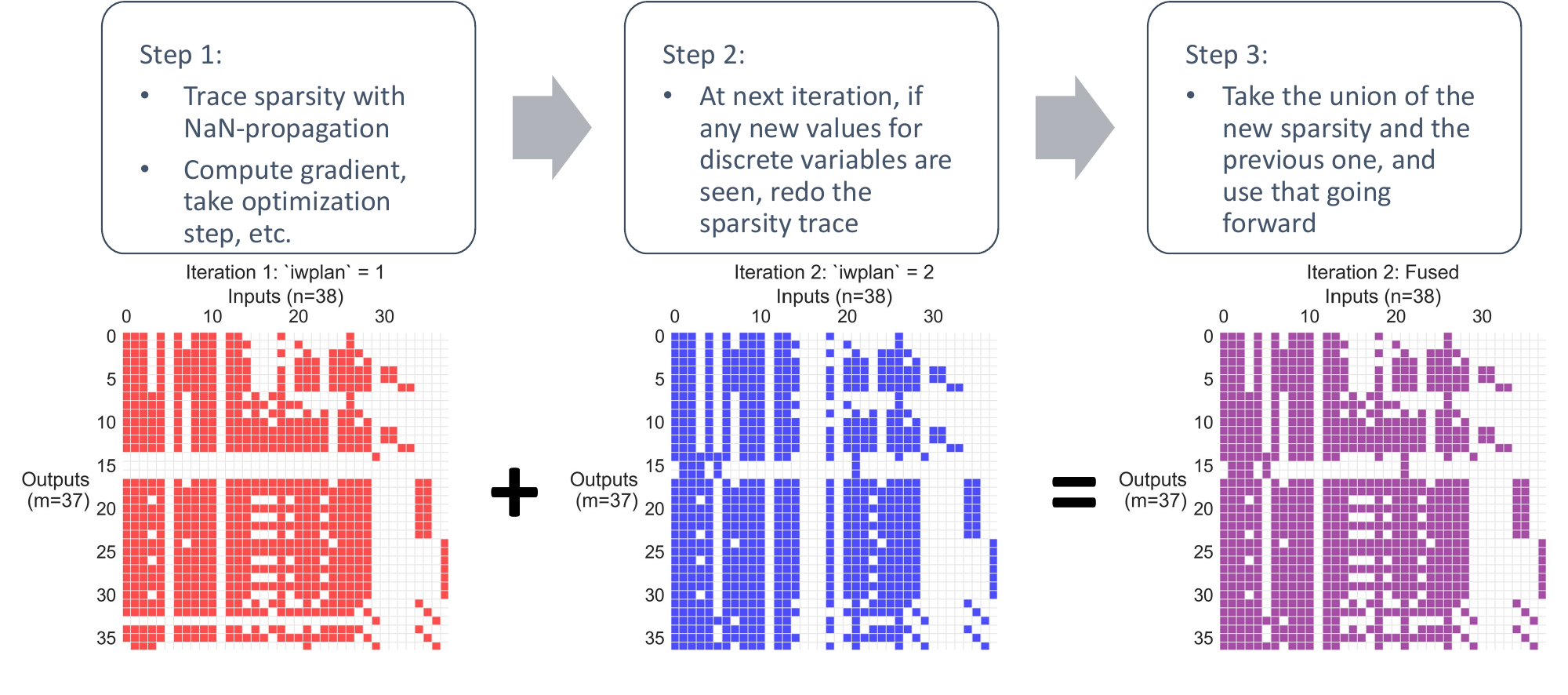}
    \caption{Illustration of a greedy algorithm for iteratively estimating the sparsity pattern of black-box models, in the presence of branching code execution. This approach is intended for use with engineering analysis models, and the process is applied to the example problem of Section \ref{sec:nan-demo}.}
    \label{fig:nan-bce-greedy}
\end{figure}

A visual illustration of this method is given in Figure \ref{fig:nan-bce-greedy}. Here, the sparsity pattern is iteratively refined as new values are observed on the discrete-typed input.

It should be emphasized that this algorithm is only a heuristic, and piecewise branching on continuous variables remains problematic. However, this strategy still offers substantial practical improvements over existing techniques that use a sparsity pattern purely based on Jacobian evaluation at the initial guess.

\section{Comparison with Existing Methods}
\label{sec:nan-comparison}

\begin{figure}[H]
    \centering
    \begin{subfigure}{0.49\textwidth}
        \centering
        \includegraphics[page=10, width=\textwidth]{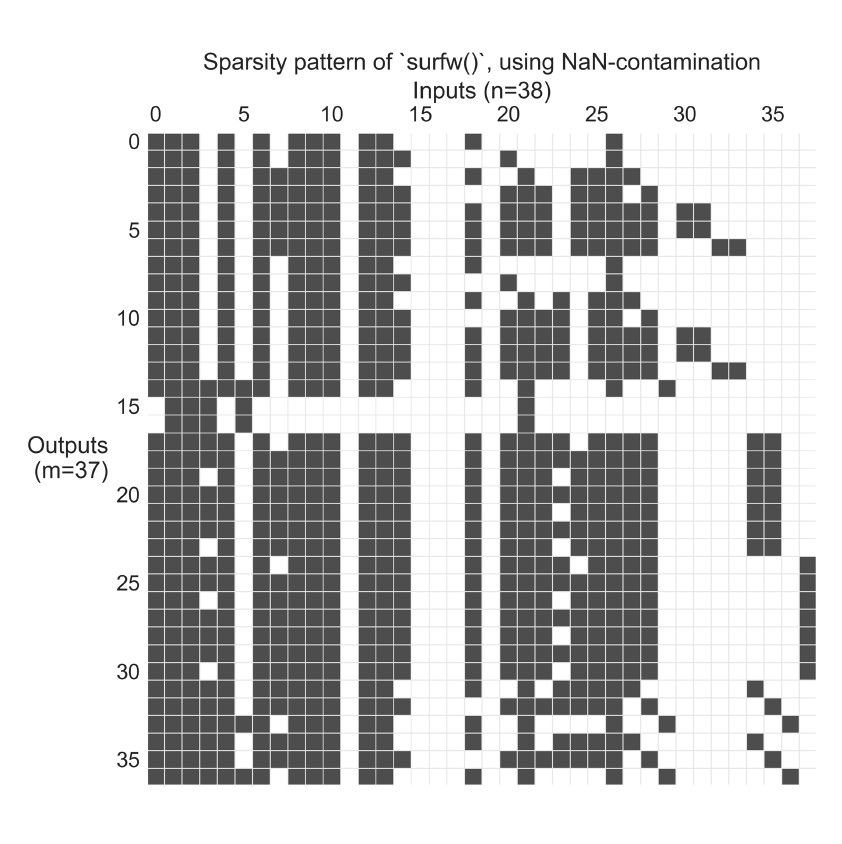}
        \caption{Sparsity pattern obtained by NaN-propagation.}
    \end{subfigure}
    \begin{subfigure}{0.49\textwidth}
        \centering
        \includegraphics[page=11, width=\textwidth]{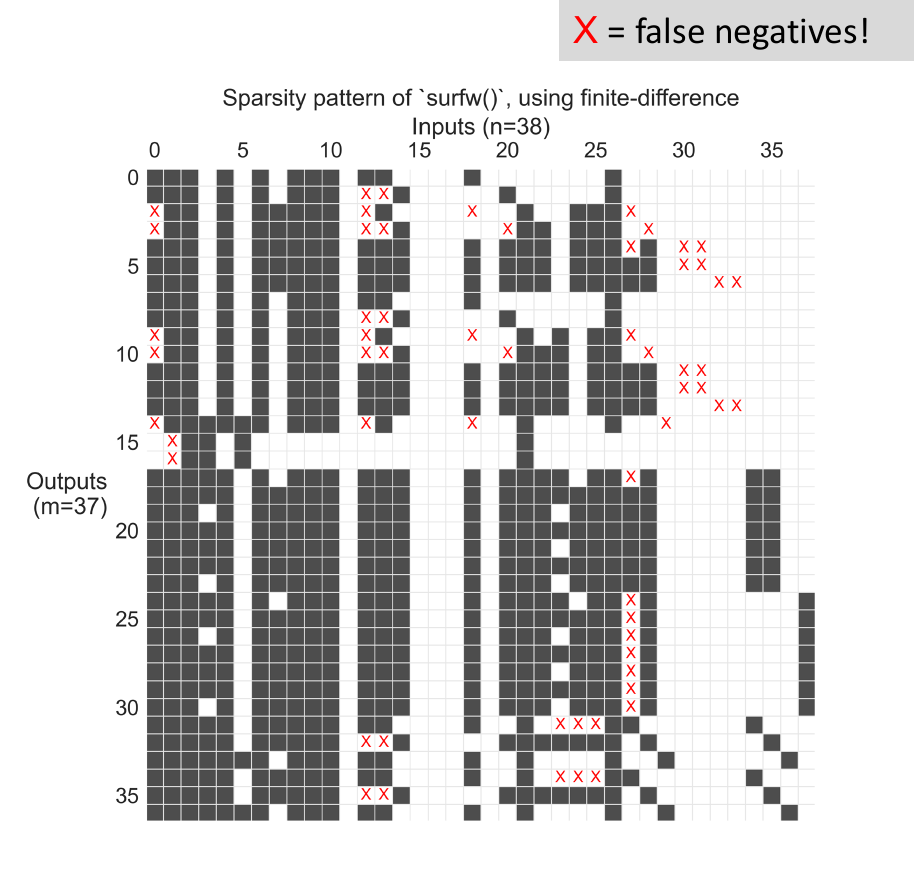}
        \caption{Sparsity pattern obtained using existing finite-difference methods. All red \textcolor{red}{\texttt{X}} marks denote false negatives that are not captured by existing methods, but detected with NaN-propagation.}
        \label{fig:nan-comparison-fd}
    \end{subfigure}
    \caption{Comparison of sparsity patterns obtained by NaN-propagation and existing finite-difference methods for the TASOPT wing weight model. Existing methods lead to false negatives via coincidental zero gradients, which can lead to costly mistakes during Jacobian compression.}
    \label{fig:nan-comparison}
\end{figure}

In summary, the main advantage of the NaN-propagation technique is that it eliminates a key false-negative failure mode of existing gradient-based sparsity detection methods. To compare this more precisely, we can show a side-by-side comparison of the sparsity patterns obtained by each method for the TASOPT wing weight model. This is given in Figure \ref{fig:nan-comparison}. In this example, the initial guess used for Jacobian evaluation is taken from the TASOPT-supplied default for aircraft design optimization. Subfigure \ref{fig:nan-comparison-fd} reveals dozens of false negatives (marked as \textcolor{red}{\texttt{X}}) that are not captured by existing methods, but are detected with NaN-propagation. This clearly demonstrates the potential utility of NaN-propagation for sparsity-tracing black-box code functions.

\section{Advanced Algorithmic Strategies}
\label{sec:nan-advanced}

This paper has introduced the core NaN-propagation technique, along with its benefits over existing methods and potential limitations. In this section, we introduce several advanced strategies that can be used to further improve the performance of NaN-propagation-based sparsity tracing.

\subsection{NaN Payload Encoding}

One possible extension is based on a recognition that not all NaN values are identical, and there are multiple unique binary representations that are interpreted as NaN values. Careful use of these values can essentially allow the encoding of additional information to aid sparsity tracing. To illustrate this, consider the binary representation of a single-precision (32-bit) floating-point NaN value, following the IEEE 754 standard:

\newcommand{\pbit}[1]{\texttt{\textcolor{mydodgerblue}{#1}}} 

\begin{equation*}
    \texttt{NaN = \textcolor{mydodgerblue}{x}111 1111 1\textcolor{mydodgerblue}{xxx xxxx xxxx xxxx xxxx xxxx}}
\end{equation*}

\begin{eqexpl}[20mm]
    \item {\texttt{0, 1}} bit values
    \item {\pbit{x}} \emph{any} bit (with the edge-case exception that not all trailing bits can be zero, or the value is interpreted as $\infty$)
\end{eqexpl}

Regardless of the value of the \pbit{x} bits, this value will be interpreted as a NaN. Therefore, single-precision NaN representations allow us to encode nearly 22 bits of information (as there are $2^{22} - 2$ unique NaN values) into these \pbit{x} bits, which we call the \emph{payload}. In double-precision (64-bit) floating point arithmetic, which is more common in engineering analysis, this payload is even larger at nearly 51 bits.

Crucially, these unique NaN payloads are propagated through mathematical operations. Therefore, we can ``tag'' individual inputs with unique NaN payloads, and observe which payloads are present in the outputs. Because NaN values are no longer fungible, \emph{multiple columns of the sparsity pattern can be simultaneously traced within a single function evaluation}. This presents a novel and fundamental theoretical improvement in the time-complexity of black-box sparsity tracing, since the number of function evaluations can now be independent of the number of inputs.

An immediate natural follow-up question to ask is how NaN values are propagated through dyadic operations: in cases where a function receives multiple payload-encoded NaN arguments, it is not obvious what the payload of the output will be. In general, most programming languages and math libraries will simply propagate one of the two arguments unmodified, and ignore the other. However, which argument is preferred varies. For example, in Python, the addition, subtraction, and multiplication operators propagate the first NaN input, while the division and exponentiation operators propagate the second NaN input.

Because of this effect, care must be taken when using NaN payload encoding. To demonstrate this, consider a dyadic black-box function of the form $f(x_1, x_2)$. Suppose we perform a sparsity trace by evaluating the function at $x_1=\texttt{NaN}_1$, $x_2=\texttt{NaN}_2$, with subscripts representing that these are uniquely identifiable $\texttt{NaN}$ values. If the output of function $f$ is $\texttt{NaN}_1$, this indicates that $f$ is surely a function of $x_1$, but the sparsity with respect to $x_2$ is unknown. Likewise, the opposite is true if the output is $\texttt{NaN}_2$. In such cases, these simultaneous column evaluations of the sparsity pattern can be thought to ``shadow'' one another: a sparsity ``hit'' on one column precludes receiving any information about the other column(s). Because of this \emph{shadowing effect}, the time complexity of NaN-contamination with payload encoding is not quite $\order(1)$, though it can still be better than the $\order(N)$ of existing approaches (discussed later). Regardless, the real benefit associated with payload encoding occurs when the output $f$ is a function of neither input, as this allows exclusion of both inputs from the sparsity pattern with a single function evaluation. This provides theoretical upper- and lower-bound cases on the method's time complexity. To reason more intuitively and loosely, any strategy that allows us to gain information about multiple inputs from a single function evaluation should be as least as fast as standard $\order(N)$ time, and often faster.

This sparsity pattern construction process via NaN-payload-encoding requires an algorithm to select which combination of columns should be simultaneously evaluated next for sparsity tracing. Development of an efficient column selection algorithm is left as an area of future work. While this may at first appear to be a variant of a vertex coloring problem, the shadowing effect creates meaningful differences and a far more fascinating algorithmic challenge. To aid such future efforts, we offer some initial observations that may guide this algorithmic development for the interested reader:

\begin{enumerate}
    \item During sparsity tracing, the state of the sparsity pattern needs to be stored in a way that captures all currently-known information: one possible representation is a trinary matrix with three possible states: \texttt{0} (no dependency), \texttt{1} (dependency), and \texttt{2} (unknown). (Other state representations that admit this partial observability exist\footnote{An example would be a continuous belief state based on priors about how engineering code sparsity patterns tend to be arranged, like block-diagonality. This is more complicated, but could allow inference based on structural patterns in the partially-observed sparsity matrix. This may also have benefits in cases where branching code execution is observed, as uncertainty about sparsity can be recorded into the state and used to strategize.}.) Because decisions about column selection need to be made while the sparsity pattern is only partially observable, any good algorithm is inherently probabilistic. This unknown factor is also why this algorithm is not simply a degenerate case of a graph coloring algorithm.
    \item A good algorithm likely benefits from some tracker of the belief state about the overall density of the sparsity pattern (i.e., the fraction of unknown entries that are likely to be dependent vs. independent). This is because the number of columns that should be evaluated simultaneously is conditional on the estimated density of the sparsity pattern. If we believe that the sparsity pattern is relatively dense, then fewer columns should be evaluated simultaneously, as the shadowing effect will be significant. Conversely, if the sparsity pattern is believed to be sparse, then we should be more aggressive about evaluating more columns simultaneously, since we will gain more information per function evaluation in expectation.

          One possible way to implement this density belief state is to use a Bayesian approach, where the belief state is updated at every step based on the results of the previous function evaluations. The initial belief state could be an uninformative prior (e.g., a uniform distribution), or one could accept a user-supplied estimate. This approach has some flaws (e.g., assuming the sparsity of each entry is independent), but it presents a reasonable starting point.

    \item An algorithm that selects which combination of columns should be evaluated next is inherently a combinatorial optimization problem. These are notoriously slow to solve, in the general case. However, if the black-box function is slow to evaluate (as it often is, in the case of higher-fidelity engineering analysis), it may be well worth it to solve this column-selection problem in order to minimize the number of required evaluations. In addition, there are strategies that can accelerate this combinatorial optimization problem in this context. These could include dynamic programming (since similar versions of this problem are solved sequentially), heuristic methods (e.g., simulated annealing, genetic algorithms), or continuous relaxation.

          As a practical matter, one possible objective function of this optimization formulation is to maximize the expected value of information gain (measured by entropy) in the sparsity pattern, given the current belief state. In principle, this is similar to some forms of Bayesian optimization.

    \item We hypothesize that best-case\footnote{ignoring the trivial case where the pattern is all-zeros and this is suspected in advance} time complexity should approach $\order(\log(N))$, as measured by the number of function evaluations and where $N$ is the number of inputs. This best-case complexity occurs in the limit of low-density sparsity patterns where the shadowing effect becomes insignificant, as the column selection procedure becomes amenable to recursive halving. A worst-case time complexity is $\order(N)$, which would occur in the case of a fully-dense sparsity pattern -- this is identical to both existing methods and NaN-propagation without payload encoding.
\end{enumerate}

\subsection{Chunking}

Another novel proposed acceleration, which we term ``chunking'', contaminates multiple adjacent elements of the input vector with NaN values simultaneously. This is similar to the multiple seeding associated with payload NaN encoding, but its inclusion here is meant to emphasize that some benefits can be achieved even without uniquely identifying NaN values, in cases where this direct bit-manipulation may not be possible or desirable.

As an example: if pairs of adjacent inputs are seeded, this has the benefit of halving the number of function calls, at the detriment of providing overly-conservative sparsity information\footnote{since one cannot determine which NaN input caused a given output to return NaN, so neither input can be ruled out -- both inputs must be assumed to be the possible cause} (i.e., one with false positives). This exploits the fact that a conservative sparsity pattern can still provide some speedup during Jacobian compression, even if it is less than what might be achieved with a more accurate sparsity pattern.

It also exploits the fact that sparsity patterns in human-written code are not random. In particular, these sparsity patterns tend to have a high degree of locality (e.g., inputs at indices 3 and 4 are much more likely to share a similar sparsity pattern than inputs at indices 3 and 30). This locality is due to the fact that, in an engineering analysis with multiple submodules, humans tend to code submodule A in its entirety before implementing submodule B. Because of this, these chunked sparsity patterns using adjacent inputs can be surprisingly accurate approximations in practice.

Thus, chunking may offer significant speedups in engineering practice by exploiting typical problem structure; a computational analogy would be how an $A^*$ graph traversal algorithm outperforms Dijkstra's algorithm in practice despite identical worst-case complexity. This and other heuristics can significantly reduce the number of black-box function evaluations required to trace the sparsity pattern.

\section{Computational Reproducibility}
\label{sec:nan-reproducibility}

All source code used to generate example results in this publication is publicly available at \url{https://github.com/peterdsharpe/nan-propagation}.

\section{Conclusions}
\label{sec:conclusions}

NaN-propagation exploits the universal contamination property of IEEE 754 NaN values to detect sparsity in black-box functions. By systematically contaminating inputs with NaN and observing which outputs become NaN, the method eliminates the coincidental zero gradient failure mode that affects existing finite-difference approaches.

The technique offers several key advantages. Most importantly, it provides conservative sparsity estimates that eliminate false negatives -- a critical improvement, since incorrect sparsity patterns can silently corrupt gradient calculations during Jacobian compression. On the TASOPT wing weight model, NaN-propagation detected dozens of dependencies missed by conventional methods and achieved a 1.52× gradient computation speedup.

The method has limitations. It can introduce false positives in functions with mathematical self-cancellation (e.g., $f(x) = x - x$), and some black-box functions may not be compatible due to internal NaN handling. Fortunately, false positives have relatively low impact, resulting in a speed penalty but without compromising numerical correctness of gradients. Like existing methods, NaN-propagation struggles with branching code execution, though the proposed greedy algorithm provides practical improvements for engineering applications.

Advanced strategies can further improve performance. NaN payload encoding allows multiple columns to be traced simultaneously, offering faster-than-linear time complexity for sparse problems. Chunking techniques exploit the locality common in engineering code to reduce function evaluations with minimal accuracy loss, again yielding speed improvements.

The universal nature of IEEE 754 NaN behavior makes this approach broadly applicable across programming languages and computational environments. This is particularly valuable for engineering optimization, where legacy and proprietary analysis codes are common and gradient-based methods are essential for high-dimensional problems.

Future work should focus on developing optimal column selection algorithms for NaN payload encoding, creating hybrid approaches that combine multiple sparsity detection methods, and implementing the technique in popular optimization and machine learning frameworks to enable widespread adoption.

\bibliographystyle{iclr2026_conference}
\bibliography{C:/Users/psharpe/library, references}

\begin{thebibliography}{17}
\providecommand{\natexlab}[1]{#1}
\providecommand{\url}[1]{\texttt{#1}}
\expandafter\ifx\csname urlstyle\endcsname\relax
  \providecommand{\doi}[1]{doi: #1}\else
  \providecommand{\doi}{doi: \begingroup \urlstyle{rm}\Url}\fi

\bibitem[Andersson et~al.(2019)Andersson, Gillis, Horn, Rawlings, and Diehl]{anderssonCasADiSoftwareFramework2019}
Joel A.~E. Andersson, Joris Gillis, Greg Horn, James~B. Rawlings, and Moritz Diehl.
\newblock {{CasADi}}: A software framework for nonlinear optimization and optimal control.
\newblock \emph{Mathematical Programming Computation}, 11\penalty0 (1):\penalty0 1--36, March 2019.
\newblock ISSN 1867-2949, 1867-2957.
\newblock \doi{10.1007/s12532-018-0139-4}.

\bibitem[Drela(1999)]{drelaIntegratedSimulationModel1999}
Mark Drela.
\newblock Integrated simulation model for preliminary aerodynamic, structural, and control-law design of aircraft.
\newblock In \emph{40th {{Structures}}, {{Structural Dynamics}}, and {{Materials Conference}} and {{Exhibit}}}, St. Louis,MO,U.S.A., April 1999. {American Institute of Aeronautics and Astronautics}.
\newblock \doi{10.2514/6.1999-1394}.

\bibitem[Drela(2010)]{drelaTASOPTTransportAircraft2010}
Mark Drela.
\newblock {{TASOPT}}: {{Transport Aircraft System Optimization}}. {{Technical Description}}, March 2010.

\bibitem[Gebremedhin et~al.(2009)Gebremedhin, Tarafdar, Pothen, and Walther]{gebremedhinEfficientComputationSparse2009}
Assefaw~H. Gebremedhin, Arijit Tarafdar, Alex Pothen, and Andrea Walther.
\newblock Efficient {{Computation}} of {{Sparse Hessians Using Coloring}} and {{Automatic Differentiation}}.
\newblock \emph{INFORMS Journal on Computing}, 21\penalty0 (2):\penalty0 209--223, May 2009.
\newblock ISSN 1091-9856, 1526-5528.
\newblock \doi{10.1287/ijoc.1080.0286}.

\bibitem[Gebremedhin et~al.(2005)Gebremedhin, Manne, and Pothen]{gebremedhinWhatColorYour2005}
Assefaw~Hadish Gebremedhin, Fredrik Manne, and Alex Pothen.
\newblock What {{Color Is Your Jacobian}}? {{Graph Coloring}} for {{Computing Derivatives}}.
\newblock \emph{SIAM Review}, 47\penalty0 (4):\penalty0 629--705, January 2005.
\newblock ISSN 0036-1445, 1095-7200.
\newblock \doi{10.1137/S0036144504444711}.

\bibitem[Griewank(1988)]{griewankAutomaticDifferentiation1988}
Andreas Griewank.
\newblock On {{Automatic Differentiation}}.
\newblock In \emph{Mathematical {{Programming}}: {{Recent Developments}} and {{Applications}}}. Argonne National Laboratory, Argonne, Illinois, 1988.

\bibitem[Hill \& Dalle(2025)Hill and Dalle]{hillSparserBetterFaster2025}
Adrian Hill and Guillaume Dalle.
\newblock Sparser, {{Better}}, {{Faster}}, {{Stronger}}: {{Sparsity Detection}} for {{Efficient Automatic Differentiation}}, June 2025.

\bibitem[Kelly(2017{\natexlab{a}})]{kellyIntroductionTrajectoryOptimization2017}
Matthew Kelly.
\newblock An {{Introduction}} to {{Trajectory Optimization}}: {{How}} to {{Do Your Own Direct Collocation}}.
\newblock \emph{SIAM Review}, 59\penalty0 (4):\penalty0 849--904, January 2017{\natexlab{a}}.
\newblock ISSN 0036-1445, 1095-7200.
\newblock \doi{10.1137/16M1062569}.

\bibitem[Kelly(2017{\natexlab{b}})]{kellyTranscriptionMethodsTrajectory2017}
Matthew~P. Kelly.
\newblock Transcription {{Methods}} for {{Trajectory Optimization}}: A beginners tutorial.
\newblock \emph{arXiv:1707.00284 [math]}, July 2017{\natexlab{b}}.

\bibitem[Kubale(2004)]{kubaleGraphColorings2004}
Marek Kubale (ed.).
\newblock \emph{Graph Colorings}.
\newblock Number 352 in Contemporary Mathematics. American Mathematical Society, Providence, R.I, 2004.
\newblock ISBN 978-0-8218-3458-9.

\bibitem[Martins \& Ning(2021)Martins and Ning]{martinsEngineeringDesignOptimization2021}
Joaquim R. R.~A. Martins and Andrew Ning.
\newblock \emph{Engineering {{Design Optimization}}}.
\newblock Cambridge University Press, 1 edition, November 2021.
\newblock ISBN 978-1-108-98064-7 978-1-108-83341-7.
\newblock \doi{10.1017/9781108980647}.

\bibitem[Martins et~al.(2003)Martins, Sturdza, and Alonso]{martinsComplexstepDerivativeApproximation2003}
Joaquim R. R.~A. Martins, Peter Sturdza, and Juan~J. Alonso.
\newblock The complex-step derivative approximation.
\newblock \emph{ACM Transactions on Mathematical Software}, 29\penalty0 (3):\penalty0 245--262, September 2003.
\newblock ISSN 0098-3500, 1557-7295.
\newblock \doi{10.1145/838250.838251}.

\bibitem[Prakash \& Gomez(2022)Prakash and Gomez]{tasopt_jl}
Prashanth Prakash and Nicolas Gomez.
\newblock {TASOPT.jl}, 2022.
\newblock GitHub repository. \url{https://github.com/MIT-LAE/TASOPT.jl}.

\bibitem[Rackauckas(2021{\natexlab{a}})]{rackauckasEngineeringTradeOffsAutomatic2021}
Christopher Rackauckas.
\newblock Engineering {{Trade-Offs}} in {{Automatic Differentiation}}: From {{TensorFlow}} and {{PyTorch}} to {{Jax}} and {{Julia}}, December 2021{\natexlab{a}}.

\bibitem[Rackauckas(2021{\natexlab{b}})]{rackauckasGeneralizingAutomaticDifferentiation2021}
Christopher Rackauckas.
\newblock Generalizing {{Automatic Differentiation}} to {{Automatic Sparsity}}, {{Uncertainty}}, {{Stability}}, and {{Parallelism}}, March 2021{\natexlab{b}}.

\bibitem[Sharpe(2021)]{sharpeAeroSandboxDifferentiableFramework2021}
Peter~D. Sharpe.
\newblock {{AeroSandbox}}: {{A Differentiable Framework}} for {{Aircraft Design Optimization}}.
\newblock Master's thesis, Massachusetts Institute of Technology, 2021.

\bibitem[Sharpe(2024)]{sharpeAcceleratingPracticalEngineering2024}
Peter~D. Sharpe.
\newblock \emph{Accelerating {{Practical Engineering Design Optimization}} with {{Computational Graph Transformations}}}.
\newblock PhD thesis, Massachusetts Institute of Technology, 2024.

\end{thebibliography}

\end{document}